\documentclass[letterpaper, 10 pt, journal, twoside]{IEEEtran}  

\IEEEoverridecommandlockouts                              

\usepackage{graphicx} 
\usepackage{amssymb}
\usepackage{hyperref}
\usepackage{amsmath} 
\usepackage{multirow}
\usepackage[T1]{fontenc}  
\usepackage{siunitx}
\usepackage{booktabs}
\usepackage{adjustbox}
\usepackage[singlelinecheck=false]{subcaption}
\usepackage[normalem]{ulem}
\usepackage{xcolor}
\usepackage{flushend}

\usepackage{enumitem}
\setlist[itemize]{align=parleft,left=0.5pt..1em}
\usepackage{caption}
\newif\ifdoubleblind
\doubleblindfalse 

\title{
Doppler-SLAM: Doppler-Aided Radar-Inertial and LiDAR-Inertial Simultaneous Localization and Mapping}
\ifdoubleblind
  \author{}
\else
\author{Dong Wang, Hannes Haag, Daniel Casado Herraez, Stefan May, Cyrill Stachniss, and Andreas Nüchter 
\thanks{Manuscript received: April 14, 2025; Revised: May 29, 2025; Accepted: July 20, 2025. This paper was recommended for publication by Editor Javier Civera upon evaluation of the Associate Editor and Reviewers' comments. This work was in parts supported by the Federal Ministry for Economic Affairs and Climate Action (BMWK) on the basis of a decision by the German Bundestag under the grant number KK5150106RL4.}%
\thanks{D. Wang and A. Nüchter are with Julius-Maximilians-Universität Würzburg, Germany. Andreas Nüchter is also with the Zentrum für Telematik e.V., Würzburg and currently International Visiting Chair at U2IS, ENSTA, Institut Polytechnique de Paris, France. H. Haag and S. May are with Nuremberg Institute of Technology Georg Simon Ohm, Germany. D. Casado Herraez is with CARIAD SE and the University of Bonn, Germany. C. Stachniss is with the Center for Robotics, University of Bonn, and the Lamarr Institute for Machine Learning and Artificial Intelligence, Germany.}
\thanks{Digital Object Identifier (DOI): see top of this page.}} 
\fi
\begin{document}
\maketitle
\markboth{IEEE Robotics and Automation Letters. Preprint Version. Accepted July, 2025}
{Wang \MakeLowercase{\textit{et al.}}: Doppler-SLAM}

\begin{abstract}
Simultaneous localization and mapping is a critical capability for autonomous systems. Traditional SLAM approaches often rely on visual or LiDAR sensors and face significant challenges in adverse conditions such as low light or featureless environments. To overcome these limitations, we propose a novel Doppler-aided radar-inertial and LiDAR-inertial SLAM framework that leverages the complementary strengths of 4D radar, FMCW LiDAR, and inertial measurement units. Our system integrates Doppler velocity measurements and spatial data into a tightly-coupled front-end and graph optimization back-end to provide enhanced ego velocity estimation, accurate odometry, and robust mapping. We also introduce a Doppler-based scan-matching technique to improve front-end odometry in dynamic environments. In addition, our framework incorporates an innovative online extrinsic calibration mechanism, utilizing Doppler velocity and loop closure to dynamically maintain sensor alignment. Extensive evaluations on both public and proprietary datasets show that our system significantly outperforms state-of-the-art radar-SLAM and LiDAR-SLAM frameworks in terms of accuracy and robustness. To encourage further research, the code of our Doppler-SLAM and our dataset are available at: 
\ifdoubleblind
\url{https://anonymous.4open.science/r/Doppler-SLAM-9CE0}. 
\else
\url{https://github.com/Wayne-DWA/Doppler-SLAM}. 
\fi
\end{abstract} 
\begin{IEEEkeywords}
  Odometry, Mapping, Localization, SLAM, Autonomous Vehicle Navigation
\end{IEEEkeywords}

\section{INTRODUCTION}
\label{sec:intro}
\IEEEPARstart{I}{n} the pursuit of robust and reliable navigation solutions, simultaneous localization and mapping (SLAM) has emerged as a cornerstone technology that enables autonomous systems to perceive and interpret their environment while estimating their own position. Traditional SLAM methods often rely on visual or LiDAR sensors. Those modalities, however, can be susceptible to poor lighting, extreme weather, or featureless terrain. SLAM using radar sensing has the potential to increase robustness to environmental variability and operate effectively under challenging conditions.

In recent years, integrating radar and LiDAR sensors with inertial measurement units (IMUs) has improved the accuracy and reliability of SLAM systems~\cite{xu2022fast}~\cite{shan2020lio}~\cite{zhang20234dradarslam}. Doppler velocity from 4D radars and frequency-modulated continuous wave (FMCW) LiDARs provides direct motion information that, when fused with inertial data, enhances motion estimation and mapping~\cite{doer2020ekf}~\cite{zhuang20234d}. However, reliable SLAM in complex or dynamic environments remains challenging, and most methods either focus on a single sensor type or fail to fully exploit Doppler information in a unified framework.
\begin{figure}
    \centering
    \includegraphics[width=1\linewidth]{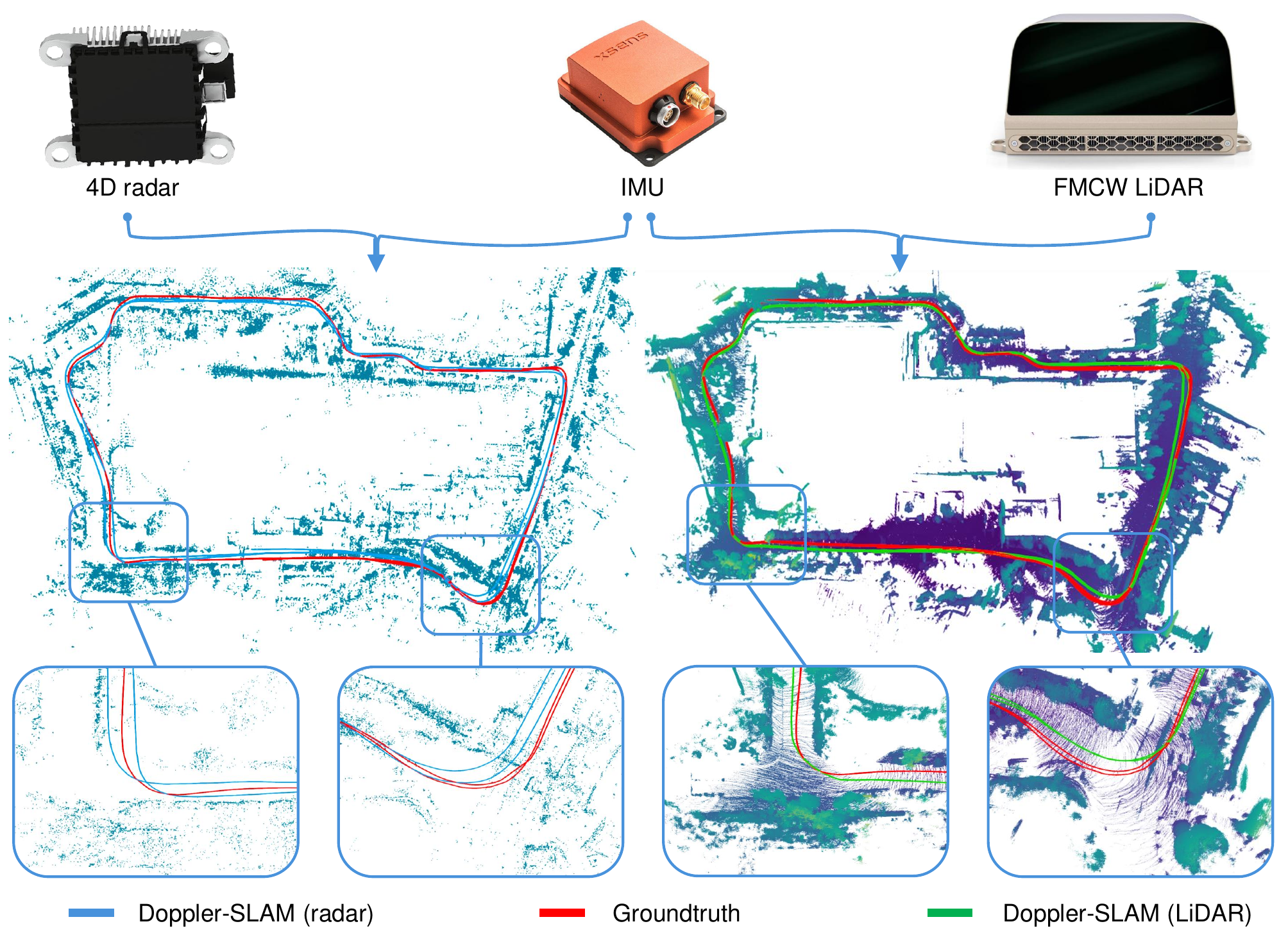}
    \caption{\small {Generalizability of our proposed Doppler-SLAM on the HeRCULES dataset "Sports Complex". Left: radar map and trajectory (\textcolor{blue}{blue}) generated with Doppler-SLAM. Right: FMCW LiDAR map and trajectory (\textcolor{green}{green}) generated with Doppler-SLAM.}} %
    \label{fig:street}
\end{figure}

This paper presents a novel approach to Doppler-aided radar-inertial and FMCW LiDAR-inertial SLAM, leveraging the complementary capabilities of radar, FMCW LiDAR, and IMU. By incorporating Doppler velocity measurements and spatial data into the SLAM framework, we aim to achieve enhanced odometry estimation and a more robust mapping process. The proposed methodology is designed to operate in complex and dynamic environments, offering a reliable solution for ground vehicles. 

The main contribution of this paper is a novel, unified SLAM approach that combines a tightly-coupled front-end~\cite{xu2022fast} with a graph optimization back-end~\cite{shan2020lio}, seamlessly integrating IMU, radar or FMCW LiDAR, and Doppler velocity measurements. Additionally, we propose an innovative online extrinsic calibration mechanism between radar-IMU or FMCW LiDAR-IMU, aided by Doppler velocity and loop closure~\cite{wang2020intensity}, to ensure consistent sensor alignment during operation, and a novel Doppler-based scan-matching method for front-end odometry, significantly improving accuracy in dynamic scenarios. Our proposed SLAM system is thoroughly evaluated on various open-source datasets alongside our dataset. The results significantly outperform current state-of-the-art radar-SLAM and FMCW LiDAR-SLAM frameworks. Finally, we make our Doppler-SLAM system open-source to foster further research and development within the community.
\section{RELATED WORK}
\label{sec:related}
In this section, we review state-of-the-art SLAM approaches based on LiDAR and radar, including traditional frameworks and recent methods that leverage Doppler information. We also discuss techniques that integrate these sensors with IMU measurements and present our proposed method in the context of Doppler-aided SLAM for radar and FMCW LiDAR. 

\textbf{LiDAR-based Odometry and SLAM} have significantly evolved with the development of various registration and optimization techniques. One of the fundamental methods for point cloud registration is the iterative closest point (ICP) algorithm, which aligns 3D shapes by minimizing point-to-point distance~\cite{besl1992method}. Generalized-ICP (GICP)~\cite{segal2009generalized} extends this by combining point-to-plane and point-to-point constraints, improving registration robustness. Building on these foundations, LOAM introduces a two-thread approach where one thread estimates motion in real-time while another refines the map~\cite{zhang2014loam}. KISS-ICP has recently been proposed as a point-to-point scan-to-map matching and keyframe-based LiDAR odometry technique that leads to high accuracy while maintaining computational efficiency~\cite{vizzo2023kiss}. However, LiDAR-only approaches struggle with featureless environments and motion distortion in highly dynamic platforms. The integration of an IMU improves the accuracy and robustness of LiDAR odometry by providing a reliable scan distortion and an initial pose for ICP. In addition, the high-frequency IMU measurements help to correct for motion distortion within LiDAR scans, which is particularly beneficial in dynamic environments where motion blur degrades scan quality. Tightly-coupling LiDAR and IMU data has also been explored through direct LiDAR-inertial fusion methods. Methods like tightly-coupled 3D LiDAR-inertial odometry~\cite{ye2019tightly} and LINS~\cite{qin2020lins} demonstrate improved state estimation accuracy through real-time optimization techniques. Graph-based approaches such as LIO-SAM also integrate LiDAR and inertial data for more accurate and globally consistent odometry~\cite{shan2020lio}. To enhance computational efficiency and real-time performance, FAST-LIO~\cite{xu2021fast} introduces a tightly-coupled iterated Kalman filter framework for robust LiDAR-inertial odometry, which is later improved with FAST-LIO2 by reducing computational complexity while maintaining accuracy~\cite{xu2022fast}. 
Recent improvements in FMCW technology have paved the way for a novel iteration of LiDAR, namely FMCW-LiDAR, which has the additional capability of measuring the relative radial velocity (Doppler velocity) of each point~\cite{wu2022picking}~\cite{papais2025balancing}. Doppler iterative closest point extends ICP by leveraging Doppler information to improve robustness in high-speed scenarios~\cite{hexsel2022dicp}. FMCW-LIO~\cite{zhao2024fmcw} and Doppler-Odom~\cite{yoon2023need} incorporate Doppler LiDAR measurements with IMU to refine motion estimation and mitigate drift in dynamic environments.

\textbf{4D-radar-based Odometry and SLAM} have gained remarkable attention due to their robustness in adverse environmental conditions, such as fog, rain, and low-light scenarios. Several approaches leverage radar-inertial fusion, probabilistic estimation techniques, and deep learning-based feature extraction to enhance odometry performance. Since 4D radar is capable of estimating the 3D ego velocity from a single scan ~\cite{kellner2013instantaneous}~\cite{wang2023infradar}, radar-only SLAM approaches can utilize the estimation of ego velocity to increase the accuracy of scan-matching ~\cite{yang2025ground}~\cite{hong2022radarslam}~\cite{park20213d}. Casado Herraez {\textit{et al.}}~\cite{herraez2024radar} propose a point-to-point ICP with Doppler velocity constraint technique specifically designed to harness the velocity information provided by radar sensors. Their approach has recently been extended to integrate IMU information, enabling fusion and optimization through global and local factor graphs~\cite{herraez2025rai}. A scan-to-submap Normal distribution transform is presented for radar point cloud registration, while velocity pre-integration is used to improve optimization performance~\cite{li20234d}. Zhang {\textit{et al.}}~\cite{zhang20234dradarslam} proposes an adaptive probability distribution-GICP to address radar measurement noise, considering the spatial probability distribution of each point in GICP \cite{segal2009generalized}. Another approach to improving the matching quality of sparse radar data is 4DiRIOM~\cite{zhuang20234d}. Here, point matching is expressed in terms of distribution-to-multiple-distribution constraints, which is achieved by matching the current scan with a sub-map constructed by the mapping module, rather than scan-to-scan matching. Huang {\textit{et al.}} \cite{huang2024less} leverage the radar cross section information to refine the point-to-point correspondence, thus improving the estimation of poses based on radar point matching. However, radar-only SLAM methods often struggle with low spatial resolution and cluttered environments, making robust feature extraction and scan-matching challenging. 
Integrating 4D radar data with inertial measurements has been shown to enhance odometry accuracy and robustness. Tightly-coupled radar-inertial odometry methods, such as DGRO~\cite{guo2024dgro}, DRIO ~\cite{chen2023drio}, and multi-state EKF-based radar-inertial odometry~\cite{michalczyk2022tightly}, integrate Doppler velocity measurements and IMU readings to provide accurate motion estimation. These methods leverage persistent landmarks and extended Kalman filtering to reduce drift in odometry estimation. Additionally, a tightly coupled factor graph formulation for radar inertial odometry ~\cite{michalczyk2024tightly} has been proposed to enhance global consistency through optimization-based approaches. Despite their advantages, radar-inertial SLAM techniques remain susceptible to sensor noise and require accurate calibration to ensure robustness.
Recent studies in deep learning and probabilistic estimation have led to novel radar odometry techniques. 
Zhou {\textit{et al.}}~\cite{zhou2022towards} leverage deep neural networks to extract robust features from radar scans for odometry estimation. Additionally, methods such as AutoPlace~\cite{cai2022autoplace} and SPR~\cite{herraez2024spr} focus on extracting point-wise features and generating scene descriptors to improve place recognition. However, these methods rely on large training datasets and may generalize poorly to unseen environments, limiting their adaptability in dynamic conditions.

Inspired by previous works~\cite{xu2022fast}~\cite{zhuang20234d}~\cite{hexsel2022dicp}~\cite{zhao2024fmcw}, we propose our Doppler-SLAM approach, which significantly enhances existing methods by unifying 4D radar-inertial and FMCW LiDAR-inertial SLAM into a single framework that directly incorporates Doppler velocity for distortion correction, scan-matching and online calibration, enabling robust SLAM in dynamic environments.
\section{Doppler-SLAM}
\label{sec:main}
To represent the various mathematical and physical quantities used in our research, we use the following conventions in this paper. Scalars will be printed as lowercase, non-bold letters (e.g., \(b\)), and constants will be printed as uppercase, non-bold letters (e.g., \(B\)). Matrices will be printed as bold upper case letters, like \(\mathbf{B}\). Vectors will be represented by bold lowercase letters, like $\mathbf{b}$. Subscripts and superscripts are used to denote different frames of reference. For example, a vector \(\mathbf{b}\) in the radar frame \(\{ \}^r\)  will be denoted as \(\mathbf{b}^r\), and the rotation from frame \(\{ \}^r\) to frame \(\{ \}^w\) will be represented by either the matrix  $\mathbf{B}_r^w$ or the quaternion \(\mathbf{b}_r^w\). The global world frame is represented by $\{ \}^W$. To simplify the exposition, the subsequent references to radar in this paper refer to 4D radar. Similarly, FMCW-LiDAR with Doppler measurements is abbreviated as LiDAR.

\subsection{Framework Overview}
\begin{figure*}[t]
        \centering
        \includegraphics[width=1\linewidth]{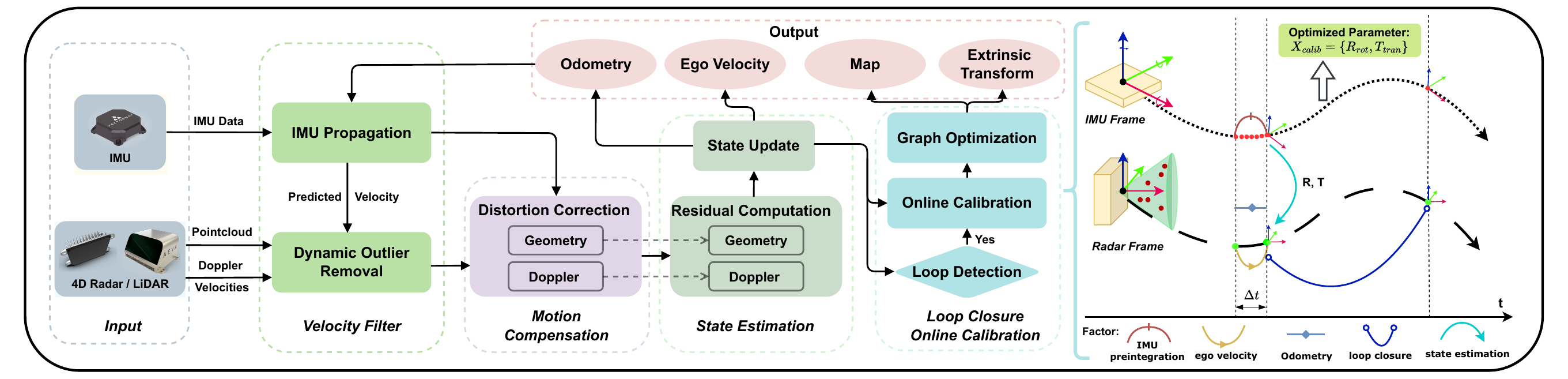}
        \caption{\small{Pipeline of Doppler-SLAM consists of four main modules: velocity filter (Sec.~\ref{velocityFilter}), motion compensation (Sec.~\ref{motionCom}), state estimation (Sec.~\ref{stateEsti}), and loop closure with graph optimization (Sec.~\ref{gtaphOpti}). The graph on the right illustrates the workflow of online extrinsic calibration between the IMU and either radar or LiDAR using graph optimization. In this approach, we combine the IMU pre-integration factor, odometry factor, and ego velocity factor to construct a factor graph. Once a loop closure factor is detected, additional optimization refines the extrinsic transformation.}}
        \label{fig:doppler_slam}
    \end{figure*}

Fig. \ref{fig:doppler_slam} illustrates the overall system architecture, highlighting four key modules: (i) velocity filter, (ii) motion compensation, (iii) state estimation, and (iv) online calibration with graph optimization. The following subsections describe the design and implementation of each module in detail.
\subsection{Front-end}
\subsubsection{\textbf{System Input and State}}
The primary input of our system is provided by an IMU and a LiDAR or a radar sensor. The measurement $\mathbf{u}$ from an IMU is defined as Eq.~\eqref{eq1}:
\begin{equation}\label{eq1}
\begin{split}
\mathbf{u} & \triangleq  \left [\hat{\boldsymbol{\omega}}_t \quad \hat{\mathbf{a}}_t  \right ],\quad
\hat{\boldsymbol{\omega}}_t =\boldsymbol{\omega}_t+\mathbf{b}_{\omega}+\mathbf{n}_{\omega}, \\
\hat{\mathbf{a}}_t & =\mathbf{R}_t^{WB}(\mathbf{a}_t-\mathbf{g})+\mathbf{b}_{\mathbf{a}}+\mathbf{n}_{\mathbf{a}},  
\end{split}\tag{1}
\end{equation}
where $\hat{\boldsymbol{\omega}}_t$ and $\hat{\mathbf{a}}_t$ are the raw IMU measurements in IMU frame \(\{ \}^B\) at time $t$. $\hat{\boldsymbol{\omega}}_t$ and $\hat{\mathbf{a}}_t$ are influenced by the slowly varying bias $\mathbf{b}$ and white noise $\mathbf{n}$. $ \mathbf{R}_t^{WB}$ is the rotation matrix from world frame \(\{ \}^W\) to IMU (body) frame \(\{ \}^B\). The term $\mathbf{g}$ refers to the constant gravity vector in the world frame. Although the measurement principles of 4D radar and FMCW-LiDAR are somewhat different, their output data is reformulated into a unified representation. Let $i$ denote the index of radar or LiDAR scans and let the points $\hat{\mathbf{m}}$ in the scan be represented as $\mathbf{s}^S_i = \{\hat{\mathbf{m}}_0,\hat{\mathbf{m}}_1, \hat{\mathbf{m}}_2, \cdots \}$, which are sampled at the local radar or LiDAR coordinate frame~$\{ \}^S$ at the end of the scan. Each point $\hat{\mathbf{m}}^S_j$ provides three-dimensional geometric point $\left [ {x, y, z}\right ]$ and radial Doppler velocity $v_j^S$ along the point's direction. The measured point~$\hat{\mathbf{m}}^S_j$ is typically affected by noise terms $\mathbf{n}^S$ and $\eta^S$, which account for geometric noise and velocity noise, respectively. Eliminating this noise recovers the true location and Doppler velocity of the point in the local sensor coordinate frame. 
\begin{equation}\label{eq2}\tag{2}
\hat{\mathbf{m}}_j^S  =\mathbf{m}_j^S + \mathbf{n}_j^S,\quad \hat{v}_j ^S =v_j^S + \eta_j^S.
\end{equation}

The system state $\mathbf{x} $ evolves on a 24-dimensional manifold and comprises the body frame’s rotation $\mathbf{R}_b $, position $\mathbf{p}_b$, and~$\mathbf{v}_b$ relative to the world frame (i.e., the initial body frame), the gyroscope and accelerometer bias $\mathbf{b}_g $ and $\mathbf{b}_a $, as well as the radar-IMU or LiDAR-IMU extrinsic parameters~$\mathbf{R}_{sb} $ and  $\mathbf{p}_{sb}$:
\begin{equation}\label{eq3}\tag{3}
\mathbf{x} \triangleq {\left [\mathbf{R}^\top_b \quad \mathbf{p}^\top_b \quad \mathbf{R}^\top_{sb}  \quad \mathbf{p}^\top_{sb}  \quad \mathbf{v}^\top_b \quad \mathbf{b}^\top_g \quad \mathbf{b}^\top_a \quad \mathbf{g}^\top \right ]}^\top.
\end{equation}
\subsubsection{\textbf{Velocity Filter}}\label{velocityFilter}
We propose a velocity filter module that fuses Doppler velocities and IMU measurements to distinguish between dynamic and static points and effectively eliminate outliers. 
\begin{figure}
    \centering
    \includegraphics[width=1\linewidth]{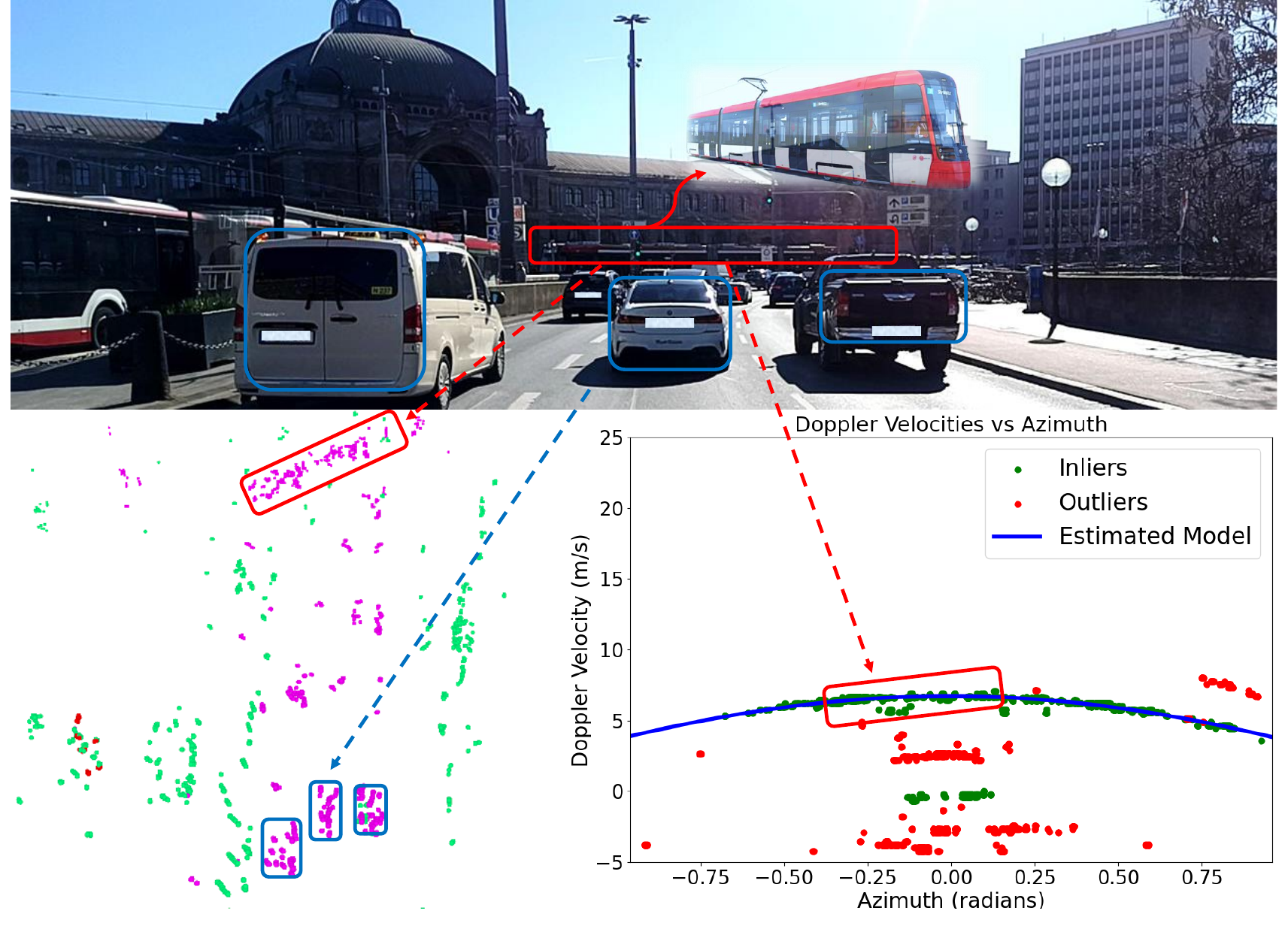}
    \caption{\small{Velocity Filter in a highly dynamic scenario with a moving tram. The top panel presents the camera view, the left panel shows the radar point cloud after processing by our proposed velocity filter, where \textcolor{purple}{purple} indicates dynamic points and \textcolor{green}{green} indicates static points, and the right panel illustrates the traditional least-squares method, in which \textcolor{green}{green} points are inliers (static objects) detected by the method and \textcolor{red}{red} points are outliers (dynamic objects). The least-squares method relies on the Doppler velocities of all inliers to fit the ego velocity profile  (\textcolor{blue}{blue} curve) but struggles in highly dynamic environments because it incorrectly incorporates Doppler measurements from moving objects (tram). In contrast, our IMU-based velocity filter effectively distinguishes between dynamic and static points, yielding more accurate ego velocity estimates and robust performance in complex, real-world scenarios.}}
    \label{fig:velocity_filter}
\end{figure}
Assume that the optimal state estimate after fusing the last sensor scan is~$\mathbf{x}_{i}$ with the covariance matrix $\boldsymbol{\xi}_{i}$. As proposed in FAST-LIO2~\cite{xu2022fast}, forward propagation starts when an IMU measurement is received and stops upon receiving a new sensor scan. 
The continuous model is discretized at the IMU sampling period~\cite{he2021embedding} based on the operation $\mathbf{\boxplus}$ and the derivative of the discrete model $\mathbf{f}$ defined in FAST-LIO \cite{xu2021fast}. Let $\Delta t$ denote the sampling period between two consecutive IMU measurements and $\mathbf{w}$ represent the noise. Then, the predicted state from the IMU is formulated as follows: 
\begin{equation}\label{eq4}\tag{4}
\begin{split}
\mathbf{x}_{i+1}  &=\mathbf{x}_{i} \boxplus\left(\mathbf{f}\left(\mathbf{x}_i, \mathbf{u}_i, \mathbf{w}_i\right) \Delta t\right) \\
\mathbf{w}_i  &\triangleq {\left [\mathbf{n}_\omega^\top \quad \mathbf{n}_a^\top \quad \mathbf{n}_{b\omega}^\top  \quad \mathbf{n}_{ba}^\top \right ]}^\top.
\end{split}
\end{equation}

At any given moment, the ego velocity is represented by~$\mathbf{v}^S$. The measured Doppler velocity $v_j^S$ from each target is considered as taking the magnitude of the projection of the relative velocity vector between the target and the sensor onto the ray connecting the target and the sensor. This calculation involves the dot product of the target's velocity $\mathbf{v}^S$ in the sensor frame $\{ \}^S$ and the unit vector pointing from the sensor towards the target:
\begin{equation}\label{eq5}\tag{5}
-v_j^S=\frac{\mathbf{m}^S}{\left\|\mathbf{m}^S\right\|} \cdot \mathbf{v}^S=\mathbf{r}^S \cdot \mathbf{v}^S=r_x^S v_x^S+r_y^S v_y^S+r_z^S v_z^S. 
\end{equation}

Considering the rigid body transformation detailed in~\cite{zhuang20234d}, the velocity $\mathbf{v}_b$ obtained from IMU forward propagation is transformed into the LiDAR or radar coordinate system $\{ \}^S$ by:
\begin{equation}\label{eq6}\tag{6}
\hat{\mathbf{v}}^S = \mathbf{R}^\top_{sb}(\mathbf{R}^\top \mathbf{v}_b + (\boldsymbol{\omega}_t - \mathbf{b}_{\omega})\times \mathbf{p}^\top_{sb}).
\end{equation}

By substituting Eq. \eqref{eq6} into Eq. \eqref{eq5}, we obtain the predicted Doppler velocity for each point:
\begin{equation}\label{eq7}\tag{7}
\hat{v}_j^S=\mathbf{r}^S \cdot \mathbf{\hat{v}}^S=r_x^S \hat{v}_x^S+r_y^S \hat{v}_y^S+r_z^S \hat{v}_z^S. 
\end{equation}

Introducing a predetermined threshold $\Upsilon$ to mitigate low-amplitude fluctuations caused by sensor noise, the velocity filter is defined by the condition $\lvert\hat{v}_j^S - v_j^S\rvert \leq \Upsilon$. This formulation ensures that a point $\hat{\mathbf{m}}^S_j$ is considered valid only if the discrepancy between its predicted Doppler velocity $\hat{v}_j^S $ and measured Doppler velocity $v_j^S$ is below $\Upsilon$, effectively filtering out dynamic outliers and ghost points. A key advantage of our proposed velocity filter over traditional least-squares methods~\cite{doer2021radar} is that it does not require the assumption that most targets in the environment are stationary, which is often an unrealistic assumption, as illustrated in Fig.~\ref{fig:velocity_filter}. This benefit is especially evident in highly dynamic outdoor environments, where least-squares approaches often fail.
\subsubsection{\textbf{Motion Compensation}}\label{motionCom}
To mitigate motion-induced distortion, our method performs a two-stage compensation of the LiDAR measurements. For radar, there is no motion distortion since its data acquisition method captures point cloud and Doppler velocities instantaneously, effectively bypassing the temporal distortions inherent in sequential LiDAR scanning.

\textbf{Geometry Compensation}: For a point $\hat{\mathbf{m}}^S_j$ sampled at timestamp $t_i^j$ in the scan $\mathbf{s}^S$ with scan-end time $t_{i}$, we adopt the backward propagation method from FAST-LIO~\cite{xu2021fast} to compensate for the geometric motion distortion. Eq.~\eqref{eq4} is backward propagated as $ \mathbf{\hat{x}}_{i-1}  =\mathbf{\hat{x}}_{i} \boxplus\left(-\mathbf{f}\left(\mathbf{\hat{x}}_i, \mathbf{u}_i, 0\right) \Delta t\right)  $. The earlier IMU measurement is used as the input for all points sampled between two consecutive IMU measurements. Subsequently, we use the relative pose between $t_i^j$ and $t_{i}$ to transform the local measurement $\hat{\mathbf{m}}^S_j$ into its corresponding scan-end measurement. In this way, the transformed points in the scan are considered to have all been scanned simultaneously at the scan-end time $t_{i}$.

\textbf{Doppler Compensation}: The measured Doppler velocity~$v_i^S$ for each point is influenced by both the target’s motion and the sensor’s motion. Dynamic targets are first filtered out by the velocity filter. Inspired by FMCW-LIO~\cite{zhao2024fmcw}, we remove the sensor’s motion over a scan period by subtracting its projected velocity change from the measured Doppler value. The relative velocity between $t_i^j$ and $t_{i}$ is also calculated by the backward propagation with IMU measurements. 

\subsubsection{\textbf{State Estimation}}\label{stateEsti}
To estimate the state vector given in Eq.~\eqref{eq3}, we employ an iterated extended Kalman filter (IEKF). The IEKF iteratively linearizes the nonlinear system around the most recent state estimate, thereby refining the estimate and enhancing the overall accuracy of the state estimation process. Keeping the first-order terms from Eq.~\eqref{eq4} and setting the noise term $\mathbf{w} $ to zero, the error state  $\mathbf{\delta x} $ and the covariance $\mathbf{\hat{P}}$ evolve according to the following linear model:
\begin{equation}\label{eq8}\tag{8}
\begin{split}
\mathbf{\delta x}_{i+1}  &=\mathbf{F}_{\delta x_i} \mathbf{\delta x}_i + \mathbf{F}_{w_i} \mathbf{w}_i,  \\
\mathbf{\hat{P}}_{i+1} &= \mathbf{F}_{\delta x_i} \mathbf{\hat{P}}_{i}\mathbf{F}_{\delta x_i}^\top + \mathbf{F}_{w_i}\mathbf{Q}_{i} \mathbf{F}_{w_i}^\top .
\end{split}
\end{equation}

Here, $\mathbf{F}_{\delta x_i} $ and $\mathbf{F}_{w_i}$ denote the transition matrix and noise Jacobian matrix, respectively, both linearized at $\mathbf{\hat{x}}_{i}$~\cite{zhao2024fmcw}. The noise covariance $\mathbf{Q}_{i}$ is obtained from IMU calibration. Assuming that the system state and covariance are denoted as $\mathbf{\hat{x}}_i$ and $\mathbf{\hat{P}}_i$ when a new scan of radar or LIDAR arrives, the iteration of the system state is described below.

\textbf{Geometry Residual}: We first project each measured point $\mathbf{m}^S_j$ in the new scan to the global frame $ \mathbf{\hat{m}}^W_j = \mathbf{\hat{T}}_{wb}\mathbf{\hat{T}}_{bs}\left(\mathbf{m}^S_j + \mathbf{n}_j^S\right)$, where $\mathbf{\hat{T}}_{wb}$ consisting of $\mathbf{R}_b$ and $\mathbf{p}_b$ represents the body frame’s pose relative to the world frame and $\mathbf{\hat{T}}_{bs} =\operatorname{inv}(\mathbf{\hat{T}}_{sb})$ represents radar-IMU or LiDAR-IMU extrinsic transform, which are all contained in the predicted state $\mathbf{\hat{x}}_i$ from Eq.~\eqref{eq8}. The five nearest neighbors of the transformed point $\mathbf{\hat{m}}^W_j$ are selected in the map using the \textit{ikd-Tree}~\cite{xu2022fast}. These neighboring points are then utilized to fit a local planar patch characterized by the normal vector~$\mathbf{u}_j$ and centroid~$\mathbf{q}^W_j$. Ideally, $\mathbf{\hat{m}}^W_j$ should lie exactly in the fitted plane. This leads to the following equation:
\begin{equation}\label{eq9}\tag{9}
\mathbf{0} = \mathbf{u}^\top_j(\mathbf{\hat{T}}_{wb}\mathbf{\hat{T}}_{bs}\left(\mathbf{m}^S_j + \mathbf{n}_j^S\right) - \mathbf{q}^W_j). 
\end{equation}

Summarizing Eq. \eqref{eq9} into a compact form and linearizing the measurement using a first-order Taylor expansion about~$\mathbf{\hat{x}}_i$ yields the following simplified form:
\begin{equation}\label{eq10}\tag{10}
\begin{split}
\mathbf{0}  &=\mathbf{^{g}\textrm{h}}_{j}\left(\mathbf{x}_i,\mathbf{m}^S_j + \mathbf{n}_j^S\right) \simeq  \mathbf{^{g}\textrm{h}}_{j}\left(\mathbf{\hat{x}}_i,0 \right) + \mathbf{^{g}\textrm{H}}^i_j\mathbf{ \widetilde{x}}_i + \mathbf{n}_i,\\
\mathbf{0}  &=\mathbf{^{g}\textrm{r}}^i_{j} + \mathbf{^{g}\textrm{H}}^i_j\mathbf{ \widetilde{x}} _i+ \mathbf{n}_i,\\
\mathbf{^{g}\textrm{r}}^i_{j} &= \mathbf{u}^T_j(\mathbf{\hat{T}}_{wb}\mathbf{\hat{T}}_{bs}\left(\mathbf{m}^S_j +  \mathbf{n}_j^S\right) - \mathbf{q}^W_j),
\end{split}
\end{equation}
where $\mathbf{x}_i =  \mathbf{\hat{x}}_i \boxplus \mathbf{ \widetilde{x}}_i$, $\mathbf{^{g}\textrm{H}}^i_j$ denotes the Jacobian matrix of geometry measurement function $\mathbf{^{g}\textrm{h}}_j$ with respect to error state $\mathbf{\widetilde{x}}_i$. Furthermore, $\mathbf{n}_i$ models the raw measurement noise associated with $ \mathbf{n}^S$ and $\mathbf{^{g}\textrm{r}}^i_{j} $ is defined as the geometry residual. 
\begin{figure}
    \centering
    \includegraphics[width=1\linewidth]{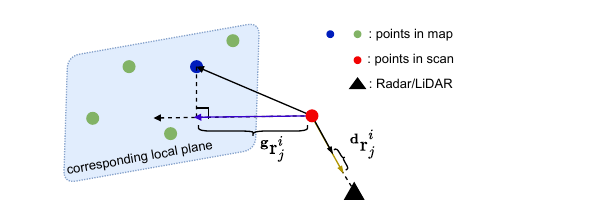}
    \caption{\small{Geometry and Doppler Residual}}
    \label{fig:residual}
\end{figure}

\textbf{Doppler Residual}: One of the key innovations of this paper is our dual matching strategy, which leverages both the observed 3D geometry and the Doppler velocity residuals for each point. The geometric residuals ensure precise alignment of the 3D point cloud, while the Doppler velocity residuals provide critical information about the motion state, enhancing overall matching accuracy. From Eq.~\eqref{eq6} and Eq.~\eqref{eq7}, we easily get the Doppler residual $\mathbf{^{d}\textrm{r}}^i_{j} $ for each point $\mathbf{m}^S_j$ in the new scan:
\begin{equation}\label{eq11}\tag{11}
\begin{split}
 \mathbf{0}  &= \sigma_i(\hat{v}_j^S -v_j^S) \simeq \mathbf{^{d}\textrm{h}}_{j}\left(\mathbf{\hat{x}}_i,0 \right) +   \mathbf{^{d}\textrm{H}}^i_j\mathbf{ \widetilde{x}}_i + \mathbf{\eta}_i,\\
\mathbf{^{d}\textrm{r}}^i_{j}  &= \mathbf{^{d}\textrm{H}}^i_j\mathbf{ \widetilde{x}}_i + \mathbf{\eta}_i \\
 &= \sigma_i(\mathbf{r}^S \mathbf{R}^\top_{sb}(\mathbf{R}^\top \mathbf{v}_b + (\boldsymbol{\omega}_t -  \mathbf{b}_{\omega})\times \mathbf{p}^\top_{sb}) -v_j^S),
\end{split}
\end{equation}
where $\sigma_i$ is the time interval between this frame and the previous one. Then, combining the prior distribution with the likelihoods derived from all geometric and Doppler observations, we obtain an equivalent maximum a posteriori~(MAP) estimate~\cite{he2021kalman} given by 
\begin{equation}\label{eq12}\tag{12}
\underset{\mathbf{ \widetilde{x}}_i}{\mathbf{min}}\left ( \left\|\mathbf{x}_i \boxminus \mathbf{ \hat{x}}_i\right\|^{2}_{\hat{\mathbf{P}}_i} +\sum ^{m}_{j=1} \left\| \mathbf{^{g}\textrm{r}}^i_{j} + \mathbf{^{g}\textrm{H}}^i_j\mathbf{ \widetilde{x}}_i + \mathbf{^{d}\textrm{r}}^i_{j} + \mathbf{^{d}\textrm{H}}^i_j\mathbf{ \widetilde{x}}_i\right\|^{2}_{\mathbf{R}_j} \right )
\end{equation}
where $\hat{\mathbf{P}}_i $, $ \mathbf{R}_j $ represent the covariance of error state and the measurement noise, respectively. $ \boxminus$ is defined in FAST-LIO~\cite{xu2021fast}. Let $\mathbf{H} = {\left [ \left (\mathbf{^{d}\textrm{H}}^i_1 + \mathbf{^{g}\textrm{H}}^i_1 \right ), \cdots, \left ( \mathbf{^{d}\textrm{H}}^i_m + \mathbf{^{g}\textrm{H}}^i_m  \right ) \right ]} ^\top$, $ \mathbf{R} = \operatorname{diag} \left (  \mathbf{R}_1, \cdots,\mathbf{R}_m \right )$, this MAP problem is solved using an IEKF with Kalman gain $\mathbf{K}$ and partial differentiation of error state $\mathbf{J}$, as below:
\begin{equation}\label{eq13}\tag{13}
\begin{split}
\mathbf{K} &=  (\mathbf{H}^\top\mathbf{R}^{-1}\mathbf{H} +\mathbf{P}^{-1})^{-1}\mathbf{H}^\top\mathbf{R}^{-1}, \\
\mathbf{P} &=  \mathbf{J}^{-1}\mathbf{\hat{P}}\mathbf{J}^{-\top}.
\end{split}
\end{equation}

Finally, after the IEKF converges, the optimal state $\mathbf{\bar{x}}_i $ and its corresponding covariance $\mathbf{\bar{P}}_i $ are given by:
\begin{equation}\label{eq14}\tag{14}
\mathbf{\bar{x}}_i =  \mathbf{\hat{x}}_i^{i+1}, \; \mathbf{\bar{P}}_i = \mathbf{I}- \left( \mathbf{K}\mathbf{H}\right)\mathbf{P}.
\end{equation}
\subsection{Graph Optimization}\label{gtaphOpti}
Online calibration and back-end optimization incorporate five principal components: IMU pre-integration, odometry, ego velocity, extrinsic transform, and loop closure factors. Among them, the IMU pre-integration, ego velocity, and extrinsic transform factors are only needed when online calibration is activated. Consequently, if the extrinsic transform between IMU and radar or LiDAR is already known, the back-end optimization is simplified to the classical loop closure optimization. The odometry and extrinsic transform factors are derived from Eq.~\eqref{eq14}, while the ego velocity factor is calculated using the least squares method combined with our proposed velocity filter. The IMU pre-integration factor connects keyframes to assist pose prediction and maintain graph constraints. Loop closure, based on ScanContext~\cite{wang2020intensity}, encodes relative poses to reduce drift, and also constrains the extrinsic estimation globally.
\section{EXPERIMENTAL EVALUATION}
\label{sec:exp}
\subsection{Hardware Setup and Dataset Collection}
\begin{figure}
    \centering
    \includegraphics[width=1\linewidth]{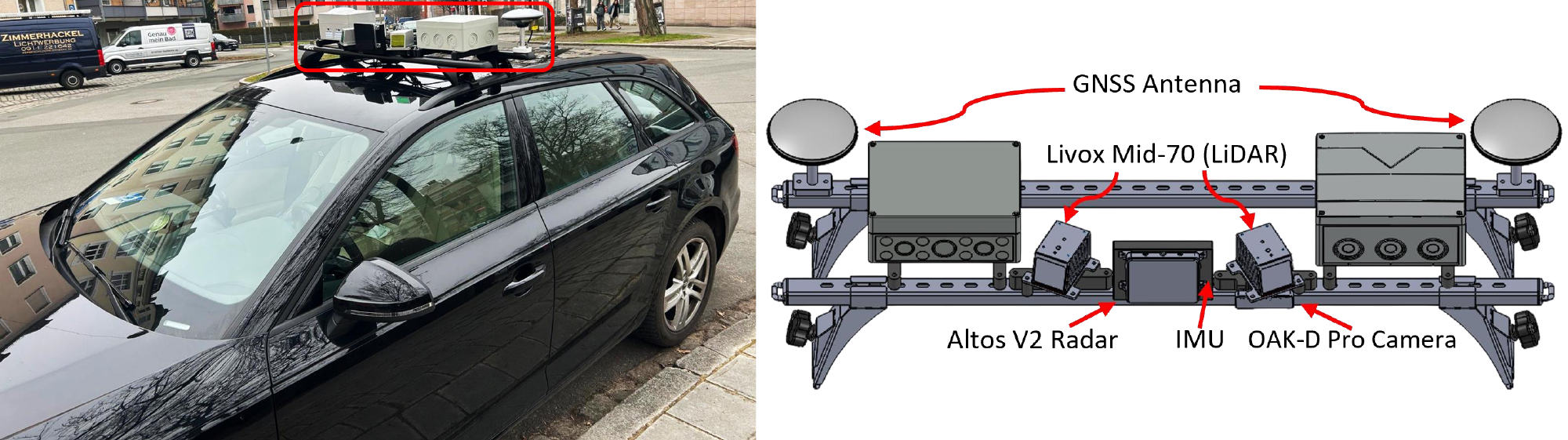}
    \caption{\small{Experiment setup (left: sensor platform mounted on a car, right: CAD Model of the platform)}}
    \label{fig:sensor_setup}
\end{figure}

As displayed in Fig.~\ref{fig:sensor_setup}, our experimental platform consists of a 4D Altos V2 radar sensor operating at 77 GHz, two Livox Mid-70 LiDAR, and a Spatial Phidget IMU. Near ground truth data was obtained using U-Blox F9 RTK-GNSS combined with inertial navigation systems, providing centimeter-level positioning accuracy and enabling precise evaluation of our method. All sensors are time-synchronized and rigidly mounted on a roof rack on top of the test vehicle to ensure accurate spatial alignment. We collect data in diverse scenarios, including urban, suburban, and highway, under varying weather and lighting conditions. LiDAR-to-IMU extrinsic calibration is performed using the method proposed in LI\_Init~\cite{zhu2022robust}, and radar-to-IMU extrinsic calibration is computed through the online calibration approach introduced in this paper.

We implement Doppler-SLAM in C++ with ROS1 and GTSAM~\cite{dellaert2012factor} and perform evaluations on a computer equipped with a 4.6 GHz AMD Ryzen 5600x CPU and 32 GB of RAM. Our evaluation metrics include absolute pose error~(APE) and relative pose error~(RPE) per frame. To thoroughly evaluate the performance of Doppler-SLAM, we conduct extensive experiments targeting accuracy, robustness, and generalization across diverse scenarios and sensor types, including various radar and LiDAR sensors. The system is benchmarked on multiple publicly available datasets, including (i) Snail-Radar~\cite{huai2024snail} using Continental ARS548 radar; (ii) NTU4DRadLM~\cite{zhang2023ntu4dradlm} employing the Oculii Eagle radar; (iii) HeRCULES~\cite{hjkim-2025-icra} using Continental ARS548 radar paired with an Aeva FMCW LiDAR; and (iv) our newly introduced IMADAR dataset with the Altos V2 radar.  
\subsection{Comparison to State-of-the-Art Methods}
In the first experiment, we analyze the performance of our system and compare it to state-of-the-art methods. The results show that our proposed Doppler-SLAM achieves superior radar-SLAM performance compared to existing methods using the Snail-Radar dataset~\cite{huai2024snail}. Next, we showcase the cross-modality generalization by comparing Doppler-SLAM with state-of-the-art radar- and LiDAR-SLAM approaches on the HeRCULES dataset~\cite{hjkim-2025-icra}, using both FMCW LiDAR and 4D radar data. All Snail-Radar and HeRCULES trajectories are evaluated in the plane, with some results adapted from~\cite{herraez2025rai}. We further validate Doppler-SLAM’s versatility on NTU4DRadLM~\cite{zhang2023ntu4dradlm}  and our IMADAR dataset, which include both vehicle-mounted and handheld scenarios. For NTU4DRadLM and IMADAR, 3D pose accuracy is assessed by evaluating the vertical direction as well, confirming Doppler-SLAM’s robustness and effectiveness in diverse and dynamic environments.

We benchmark Doppler-SLAM against several state-of-the-art methods, including 4DRadarSLAM (Radar-only SLAM)~\cite{zhang20234dradarslam}, Graph-RIO (Radar-inertial odometry)~\cite{girod2024robust}, Radar-ICP (Doppler velocity aided Radar-only odometry)~\cite{herraez2024radar}, Go-RIO (Doppler velocity aided Radar-inertial odometry)~\cite{yang2025ground}, KISS-ICP (LiDAR odometry approach on radar)~\cite{vizzo2023kiss}, FAST-LIO2 (LiDAR-inertial odometry)~\cite{xu2022fast}, and RIV-SLAM (Radar-inertial SLAM)~\cite{wang2024riv}.

Quantitative results on the Snail-Radar dataset are presented in TABLE~\ref{tab:slam_comparison_snail}. FAST-LIO2 with loop closure (FAST-LIO2-LC) on LiDAR data serves as the baseline, while all other methods operate on radar data. KISS-ICP provides good local accuracy on radar point clouds but suffers from increased drifts in large dynamic scenes. Both Radar-ICP and 4DRadarSLAM are radar-only, point-to-point matching methods that, without motion constraints, are prone to incorrect matches in large-scale or highly dynamic sequences such as 20240115/2 and 20240123/2. Radar-IMU-based methods, such as Graph-RIO and RIV-SLAM, despite using IMU as motion constraints, also exhibit significant errors due to their failure to detect loop closure in dynamic environments. Doppler-Odometry (without loop closure) outperforms all other radar-based methods by utilizing Doppler velocities to improve motion estimation. Notably, Doppler-SLAM further enhances performance by incorporating loop closure, achieving accuracy on radar data that is comparable to LiDAR-based SLAM.
\begin{table*}[ht]
\centering
\resizebox{\textwidth}{!}
{\begin{tabular}{@{}l|ccc|ccc|ccc|ccc|ccc@{}}
\toprule
\multirow{2}{*}{Method} & \multicolumn{3}{c|}{\textbf{20240113/3} (4.6 km)}       & \multicolumn{3}{c|}{\textbf{20240113/1} (0.5 km)}       & \multicolumn{3}{c|}{\textbf{20240115/2} (6.6 km)}       & \multicolumn{3}{c|}{\textbf{20240123/2} (8.5 km)}       & \multicolumn{3}{c}{\textbf{20240123/3} (2.2 km)}        \\ \cmidrule(l){2-16} 
& RPE {[}m{]}    & RPE {[}°{]}    & APE {[}m{]}  & RPE {[}m{]}    & RPE {[}°{]}    & APE {[}m{]}  & RPE {[}m{]}    & RPE {[}°{]}    & APE {[}m{]}  & RPE {[}m{]}    & RPE {[}°{]}    & APE {[}m{]}  & RPE {[}m{]}    & RPE {[}°{]}    & APE {[}m{]}  \\ \midrule
FAST-LIO2-LC (LiDAR)& \textbf{0.007} & \textbf{0.011} & \uline{1.870}& \textbf{0.010}& \textbf{0.013}& 0.359& \textbf{0.007}& \textbf{0.008}& \textbf{1.183}& \textbf{0.022}& \textbf{0.005}& \uline{9.889}& \textbf{0.009}& \textbf{0.008}& \textbf{0.693}\\ \hline
KISS-ICP (radar)                 & 0.240          & 0.155& 68.40& 0.118& 0.179& 4.389& 0.232          & 0.134          & 147.1        & 0.269& 0.117 & 167.8        & 0.222          & 0.155& 45.86\\
Radar-ICP                        & 0.238          & 0.156          & 18.24& 0.120& 0.174& 3.946& 0.229& 0.131& 31.62& 0.252& 0.112& 37.47& 0.221& 0.151& 7.893\\
4DRadarSLAM                      & 0.737          & 1.170& 53.23& 0.460& 1.074& 8.908& 0.663          & 1.179& 491.2& 0.864& 0.901& 454.5& 0.503& 0.983& 142.1\\
Graph-RIO                        & -& -& -& 0.169& 0.172& 9.523& 0.195& 0.172& 763.4& -& -& -& 0.266& 0.168& 497.0\\
RIV-SLAM                         & 0.213          & 0.142          & 30.17& 0.113& 0.171& 4.131& 0.219          & 0.128& 33.1& 0.224& 0.101& 35.51& 0.201          & 0.140          & 6.120          \\ \midrule
\textbf{Doppler-Odometry}(radar) & 0.151& 0.113& 3.375& 0.083& 0.167& \textbf{0.285}& 0.175& 0.102& 9.391& 0.199& \uline{0.095}& 10.59& \uline{0.156}& 0.117& 2.608\\
\textbf{Doppler-SLAM}(radar)     & \uline{0.150}& \uline{0.111}& \textbf{\uline{1.532}}& \uline{0.082}& \uline{0.160}& \uline{0.316}& \uline{0.174}& \uline{0.098}& \uline{5.651}& \uline{0.198}& \uline{0.095}& \textbf{\uline{5.810}}& \uline{0.156}& \uline{0.116}& \uline{1.556}\\ \bottomrule
\end{tabular}}
\caption{\small{Comparison of SLAM methods on Snail-Radar dataset. \textbf{Bold}: best results, \uline{underlined}: best radar results.}}
\label{tab:slam_comparison_snail}
\end{table*}

TABLE \ref{tab:slam_comparison_hercules} presents the evaluation of Doppler-SLAM's performance and generalizability on the HeRCULES dataset, examining both radar and FMCW LiDAR data. To highlight the improvements of Doppler-SLAM over traditional LiDAR-SLAM methods on FMCW LiDAR, we provide a comparative analysis with FAST-LIO2~\cite{xu2022fast}. Thanks to the robustness of the velocity filter in dynamic scenarios and the tight coupling between IMU measurements and Doppler velocities, Doppler-SLAM consistently outperforms FAST-LIO2 in the sequences "Street Day", characterized by highly dynamic conditions and rain, "Library Day", and "Parking Lot", notable for frequent sharp turns. These results demonstrate that Doppler-SLAM maintains remarkable robustness in harsh conditions where competing methods suffer significant performance degradation. These capabilities are further highlighted on the NTU4DRadarLM dataset, as shown in TABLE~\ref{tab:slam_ntu4d}, which employs the Oculii Eagle radar with both handheld and vehicle-mounted data acquisition. 
\begin{table*}[ht]
\centering
\resizebox{\textwidth}{!}
{\begin{tabular}{@{}l|ccc|ccc|ccc|ccc|ccc@{}}
\toprule
\multirow{2}{*}{Method} & \multicolumn{3}{c|}{\textbf{Mountain Day 1} (4 km, mountain)} & \multicolumn{3}{c|}{\textbf{Library Day 1} (1.6 km)} & \multicolumn{3}{c|}{\textbf{Sports Complex Day 1} (1.4 km)}  & \multicolumn{3}{c|}{\textbf{Parking Lot 3 Night} (0.5 km)} & \multicolumn{3}{c}{\textbf{Street Day 1} (1 km, rain)}        \\ \cmidrule(l){2-16} 
& RPE {[}m{]}    & RPE {[}°{]}    & APE {[}m{]}  & RPE {[}m{]}    & RPE {[}°{]}    & APE {[}m{]}  & RPE {[}m{]}    & RPE {[}°{]}    & APE {[}m{]}  & RPE {[}m{]}    & RPE {[}°{]}    & APE {[}m{]}  & RPE {[}m{]}    & RPE {[}°{]}    & APE {[}m{]}  \\ \midrule
FAST-LIO2-LC (LiDAR)          & \textbf{0.085}& 0.061& 4.595& 0.095& 0.072& 5.234& \textbf{0.084}& 0.073& 2.114& \textbf{0.091}& 0.092& 1.828& \textbf{0.028}& 0.026& 2.966\\
RIV-SLAM (radar)              & 0.077& 0.084& 206.9& 0.014 & \uline{0.064} & \uline{4.175} & -& -& -& 0.020 & \uline{0.075}& 2.380& \uline{0.010}& 0.042& 10.74\\ 
Radar-ICP (radar)             & \uline{0.055}& \uline{0.067}& 118.6& 0.049& \uline{0.064}& 10.23& \uline{0.049} & 0.071& 7.125& 0.058& 0.089& 3.414& 0.022& \uline{0.028} & 11.66\\ \midrule
\textbf{Doppler-SLAM} (radar) & 0.059& 0.068& \uline{43.05}& \uline{0.012}& 0.094& 13.24& 0.081& \uline{0.065} & \uline{2.718} & \uline{0.015}& 0.078& \uline{0.717}& 0.013& 0.039& \uline{7.448}\\ 
\textbf{Doppler-SLAM} (LiDAR) & 0.128& \textbf{0.054}& \textbf{4.498}& \textbf{0.085}& \textbf{0.061}& \textbf{3.365}& 0.093& \textbf{0.067}& \textbf{2.069}& 0.101& \textbf{0.077}& \textbf{1.642}&  0.029& \textbf{0.024}& \textbf{2.813}\\ \bottomrule
\end{tabular}}
\caption{\small{Comparison of SLAM methods on HeRCULES dataset. \textbf{Bold}: best LiDAR results, \uline{underlined}: best radar results.}}
\label{tab:slam_comparison_hercules}
\end{table*}
\begin{table}[htbp]
\centering
\resizebox{0.49\textwidth}{!}{%
\begin{tabular}{l|ccc|ccc|ccc}
\toprule
\multirow{2}{*}{Method}  & \multicolumn{3}{c|}{\textbf{cp} (handcart, 0.25 km)}& \multicolumn{3}{c|}{\textbf{loop2} (car, 4.79 km)} & \multicolumn{3}{c}{\textbf{loop3} (car, 4.23 km)}\\ \cmidrule(l){2-10}
 & RPE [m] & RPE [°] & APE [m] & RPE [m] & RPE [°] & APE [m]  & RPE [m] & RPE [°] & APE [m] \\
\midrule
4DRadarSLAM & 0.129& \uline{0.255} & \textbf{0.861}& 1.337& 0.308& \uline{43.67} & 1.216& 0.455& 334.7\\
\midrule
Go-RIO & \uline{0.079}& 0.661& \uline{1.035}& -& -& -& 0.991& 1.035& 52.74\\
\midrule
\textbf{Doppler-Odometry} & \textbf{0.022}& \textbf{0.109}& 3.267& \uline{0.260}& \uline{0.179}& 50.56& \uline{0.209}& \textbf{0.162}& \uline{48.34}\\
\midrule
\textbf{Doppler-SLAM} &0.079& 0.267& 1.508& \textbf{0.261}& \textbf{0.126}& \textbf{4.278}& \textbf{0.188}& \uline{0.169}& \textbf{8.182}\\
\bottomrule
\end{tabular}%
}
\caption{\small{Comparison of SLAM methods across sequences on NTU4DRadarLM dataset. \textbf{Bold}: best results, \uline{underlined}: second best results.}}
\label{tab:slam_ntu4d}
\end{table}

\begin{figure}
    \centering
    \includegraphics[width=1\linewidth]{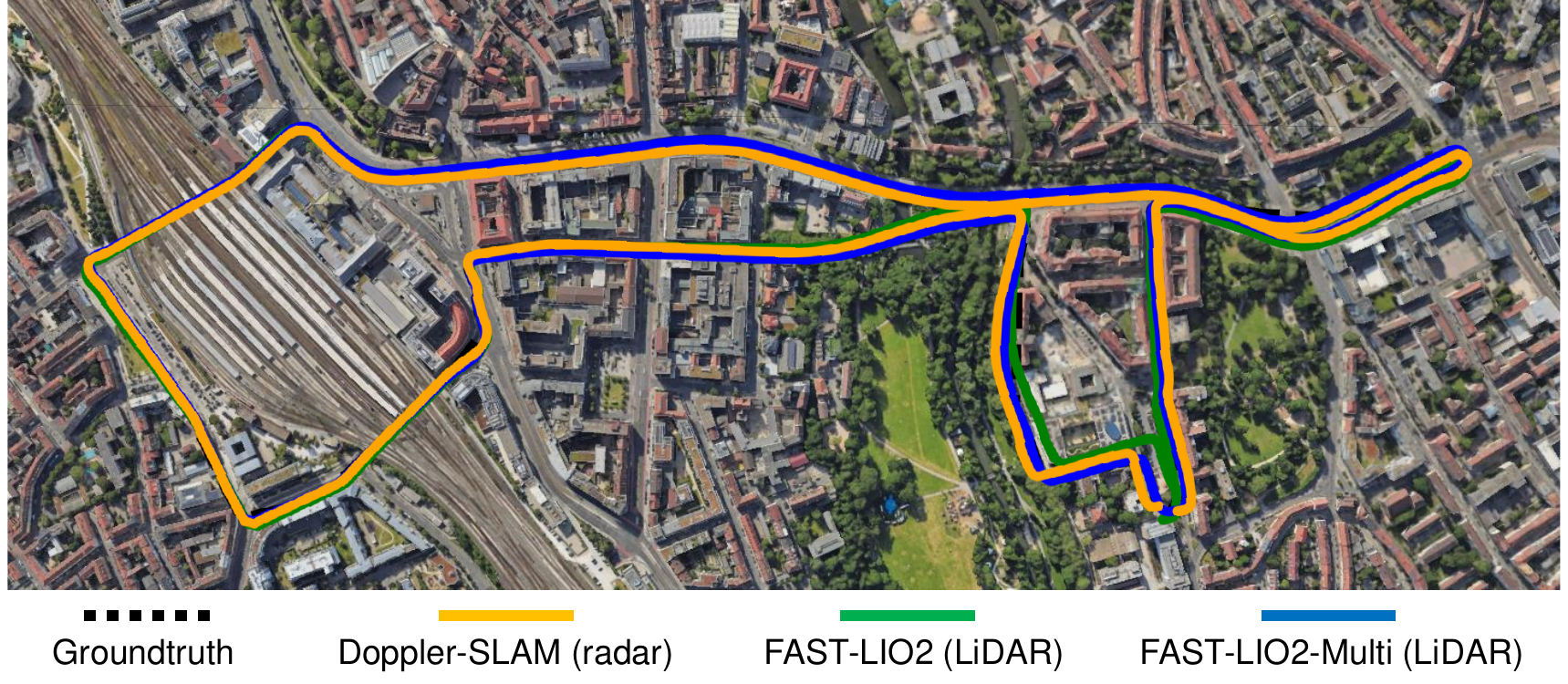}
    \caption{\small{Qualitative results on sequence "HBF" from IMADAR dataset.}}
    \label{fig:imadar_2}
\end{figure}

However, the datasets mentioned above are mostly low-speed (below 50\,km/h) driving scenarios. To more comprehensively evaluate the performance of Doppler-SLAM in high-speed, long-distance scenarios, we conduct additional experiments using our own IMADAR dataset, benchmarked against FAST-LIO2 (FAST-LIO2-Multi for two LiDARs with asynchronous update) with loop closure for comparative analysis. TABLE~\ref{tab:slam_imadar} presents the evaluation of three different sequences:~"WoehrSee" and~"HBF", both representing urban traffic scenarios where the main challenges are complex dynamic conditions (as shown in Fig.~\ref{fig:velocity_filter}) and degraded environments such as tunnels, and the~"N4" sequence, representing high-speed and long-distance conditions with vehicle speeds up to 110\,km/h. Benefiting from a higher frequency of asynchronous updates and a larger field-of-view angle, FAST-LIO-Multi performs best on all three sequences. The quantitative results indicate that although both Doppler-SLAM and FAST-LIO2 show performance degradation over the three sequences, Doppler-SLAM using radar is still comparable to state-of-the-art LiDAR approaches and even outperforms FAST-LIO with LiDAR in highly dynamic environments on sequences "HBF"~(as illustrated in Fig.~\ref{fig:imadar_2}) and "N4". These results further validate our proposed online extrinsic calibration method, as it is consistently employed for radar-to-IMU calibration across all three sequences.
\begin{figure}
    \centering
    \includegraphics[width=1\linewidth]{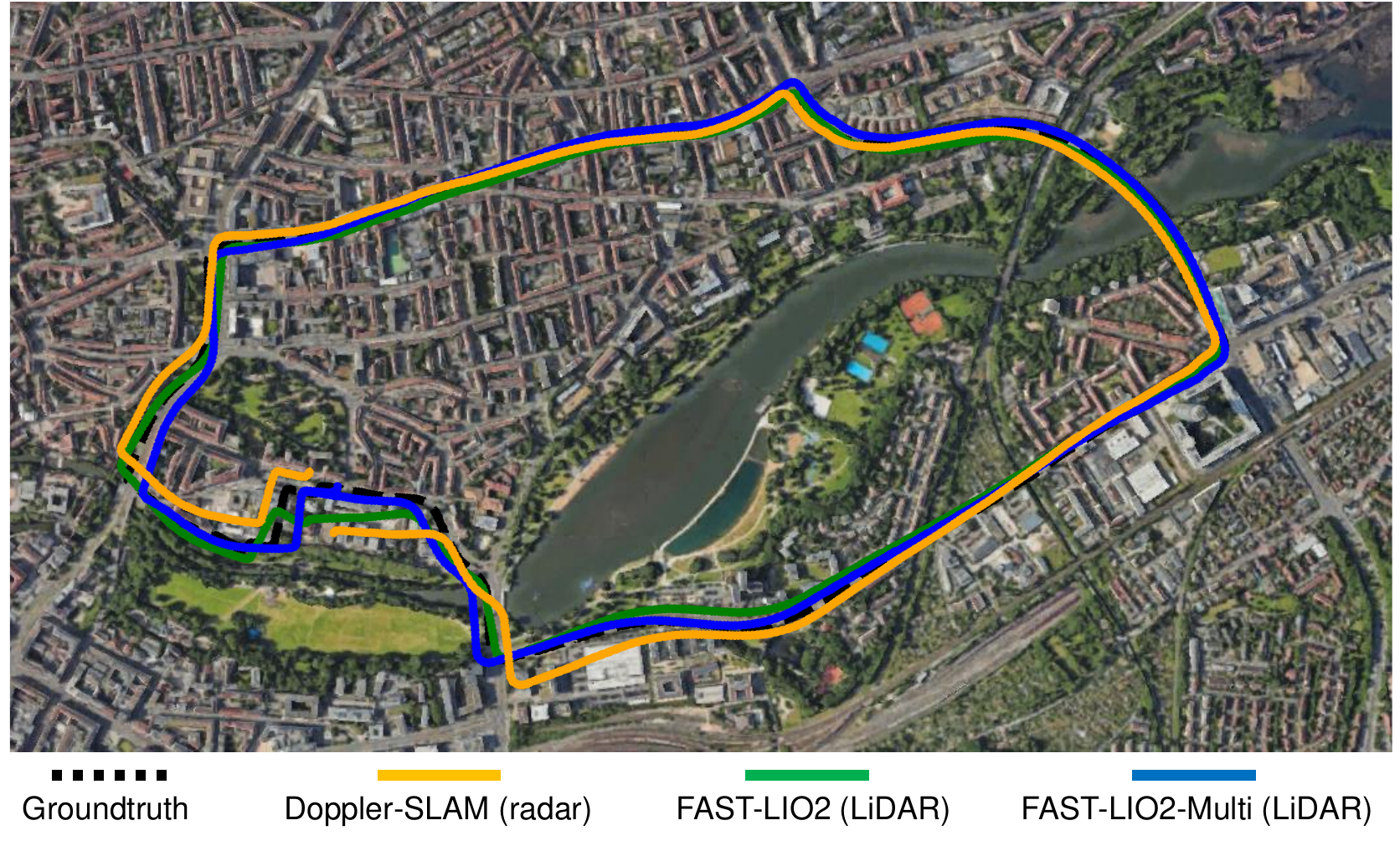}
    \caption{\small{Qualitative results on sequence "WoehrSee" from IMADAR dataset.}}
    \label{fig:imadar_1}
\end{figure}


\begin{table}[htbp]
\centering
\resizebox{0.49\textwidth}{!}{%
\begin{tabular}{l|cc|cc|cc}
\toprule
\multirow{2}{*}{Method} & \multicolumn{2}{c|}{\textbf{WoehrSee} (7.1 km)} & \multicolumn{2}{c|}{\textbf{HBF} (4.7 km)} & \multicolumn{2}{c}{\textbf{N4} (15.4 km)} \\ \cmidrule(l){2-7}
 & RPE [m] & APE [m] & RPE [m] & APE [m]  & RPE [m] & APE [m]  \\ \midrule
FAST-LIO2 (LiDAR)  & 0.22    & \uline{24.89}   & 0.29   & 15.56 & 0.252   & 138.77 \\ \midrule
FAST-LIO2-Multi (LiDAR)  & \textbf{0.115}   & \textbf{18.65}   & \textbf{0.104}    & \textbf{2.50} & \textbf{0.153}   & \textbf{58.75} \\ \midrule
\textbf{Doppler-SLAM (radar) }   & \uline{0.178}  & 30.55 & \uline{0.21}  & \uline{8.46}  & \uline{0.216}  & \uline{102.9} \\ \bottomrule
\end{tabular}%
}
\caption{\small{Comparison of SLAM methods across sequences on IMADAR dataset. \textbf{Bold}: best results, \uline{underlined}: second best results.}}
\label{tab:slam_imadar}
\end{table}
\subsection{Ablation Studies}
We perform ablation studies to evaluate the contributions and computation time of individual components within the Doppler-SLAM framework. To isolate and eliminate sensor-related effects, we select sequence 20240113/3 from the Snail-Radar for the radar-based Doppler-SLAM and the sequence "Street Day" from HeRCULES for FMCW LiDAR-based ablation study. Both sequences represent dynamic and complex environments. As presented in TABLE~\ref{tab:ablation}, the results clearly demonstrate that the integration of Doppler velocity into both radar and FMCW LiDAR systems greatly improves accuracy and robustness by reducing false matches in dynamic scenes. The computation time for each module is summarized in TABLE~\ref{tab:timecomputation}. The velocity filter and online calibration modules are highly efficient, each contributing less than 0.2\,ms per frame. Doppler residual computation is also lightweight, with slightly higher cost on LiDAR due to increased point cloud density. As expected, loop closure is the most computationally intensive component, but it remains within real-time capability for both sensor types. This confirms the efficiency and suitability of Doppler-SLAM for real-time radar and FMCW LiDAR applications.
\begin{table}[htbp]
\centering
\resizebox{0.49\textwidth}{!}{%
\begin{tabular}{l|ccc|ccc}
\toprule
\multirow{2}{*}{Method} & \multicolumn{3}{c|}{\textbf{20240113/3} (radar)} & \multicolumn{3}{c}{\textbf{Street Day} (FMCW LiDAR)} \\ \cmidrule(l){2-7}
 & RPE [m] & RPE [°] & APE [m] & RPE [m] & RPE [°] & APE [m]  \\ \midrule
 w/o velocity filter& 0.216&0.3730& 16.07& 0.112& 0.102& 11.06\\ \midrule
 w/o Doppler residual & \textbf{0.149}& \uline{0.1111}& 5.671& 0.071& 0.080& 3.350\\ \midrule
  w/o online calibration& 0.159& 0.1112& 4.140& 0.052& \uline{0.038}&3.476\\ \midrule 
 w/o loop closure & 0.151& 0.113& \uline{3.378}&  \uline{0.046}& 0.039&	\uline{2.940}\\ \midrule 
\textbf{ Doppler-SLAM}  & \uline{0.150}& \textbf{0.1110}& \textbf{1.532}& \textbf{0.029}& \textbf{0.024}& \textbf{2.813}\\ \bottomrule
\end{tabular}%
}
\caption{\small{Ablation evaluation on radar and FMCW LiDAR sequences. \textbf{Bold}: best results, \uline{underlined}: second best results.}}
\label{tab:ablation}
\end{table}

\begin{table}[htbp]
\centering
\resizebox{0.49\textwidth}{!}{%
\begin{tabular}{l|c|c|c|c}
\toprule
                       & velocity filter & Doppler residual & online calibration         & loop closure   \\ \midrule
 20240113/3 (radar)    & 0.024& 1.877& 0.036& 37.72\\ \midrule
 Street Day (FMCW LiDAR)  & 0.175& 10.8& 0.097& 137\\ \bottomrule
\end{tabular}%
}
\caption{\small{Computation time of each module (ms).}}
\label{tab:timecomputation}
\end{table}

\section{CONCLUSIONS}
\label{sec:conclusion}
This paper proposes Doppler-SLAM, a novel Doppler-aided radar-inertial and LiDAR-inertial SLAM framework. By incorporating Doppler velocities into scan-matching, our approach unifies FMCW LiDAR- and 4D radar-based SLAM systems into a single framework, enabling robust SLAM performance under challenging dynamic conditions. This tightly integrated system fuses IMU data with either 4D radar or FMCW LiDAR to deliver odometry, ego velocity estimation, mapping, and extrinsic calibration between the IMU and the radar or LiDAR sensor. Our innovative online calibration technique, enhanced by Doppler velocity and loop closure, ensures consistent sensor alignment. Thorough evaluations demonstrate clear advantages over existing state-of-the-art radar-SLAM and FMCW LiDAR-SLAM frameworks, and by releasing Doppler-SLAM and our IMADAR dataset as open-source software, we encourage continued advancement and further research within the SLAM community.
\bibliographystyle{IEEEtran}
\bibliography{bibliography.bib}

@STRING{arxiv                  = {arXiv preprint} }

@STRING{case                   = {Proc.~of the Intl.~Conf.~on Automation Science and Engineering (CASE)} }

@STRING{icra                   = {Proc.~of the IEEE Intl.~Conf.~on Robotics \& Automation (ICRA)} }

@STRING{ijrr                   = {Intl.~Journal~of Robotics Research (IJRR)} }

@STRING{iros                   = {Proc.~of the IEEE/RSJ Intl.~Conf.~on Intelligent Robots and Systems (IROS)} }

@STRING{itsc                   = {Proc.~of the IEEE Intl.~Conf.~on Intelligent Transportation Systems (ITSC)} }

@STRING{MFI                    = {Proc.~of the Intl.~Conf.~on Multisensor Fusion and Integration for Intelligent Systems (MFI)}}

@STRING{ral                    = {IEEE Robotics and Automation Letters (RA-L)} }

@STRING{rss                    = {Proc.~of Robotics: Science and Systems (RSS)} }

@STRING{sensors                = {Sensors} }

@STRING{tro                    = {IEEE Trans.~on Robotics (TRO)} }

@inproceedings{besl1992method,
  title={Method for registration of 3-D shapes},
  author={Besl, Paul J and McKay, Neil D},
  booktitle={Sensor fusion IV: control paradigms and data structures},
  volume={1611},
  pages={586--606},
  year={1992},
}

@inproceedings{segal2009generalized,
  title={Generalized-icp.},
  author={Segal, Aleksandr and Haehnel, Dirk and Thrun, Sebastian},
  booktitle= rss,
  year={2009},
}

@inproceedings{zhang2014loam,
  title={LOAM: Lidar odometry and mapping in real-time.},
  author={Zhang, Ji and Singh, Sanjiv and others},
  booktitle=rss,
  year={2014},
}

@inproceedings{ye2019tightly,
  title={Tightly coupled 3d lidar inertial odometry and mapping},
  author={Ye, Haoyang and Chen, Yuying and Liu, Ming},
  booktitle=icra,
  year={2019},
}

@inproceedings{qin2020lins,
  title={Lins: A lidar-inertial state estimator for robust and efficient navigation},
  author={Qin, Chao and Ye, Haoyang and Pranata, Christian E and Han, Jun and Zhang, Shuyang and Liu, Ming},
  booktitle=icra,
  year={2020},
}

@article{xu2021fast,
  title={Fast-lio: A fast, robust lidar-inertial odometry package by tightly-coupled iterated kalman filter},
  author={Xu, Wei and Zhang, Fu},
  journal=ral,
  volume={6},
  number={2},
  pages={3317--3324},
  year={2021},
  publisher={IEEE}
}

@article{xu2022fast,
  title={Fast-lio2: Fast direct lidar-inertial odometry},
  author={Xu, Wei and Cai, Yixi and He, Dongjiao and Lin, Jiarong and Zhang, Fu},
  journal=tro,
  volume={38},
  number={4},
  pages={2053--2073},
  year={2022},
  publisher={IEEE}
}

@article{hexsel2022dicp,
  title={DICP: Doppler iterative closest point algorithm},
  author={Hexsel, Bruno and Vhavle, Heethesh and Chen, Yi},
  journal={arXiv preprint arXiv:2201.11944},
  year={2022}
}

@article{zhao2024fmcw,
  title={FMCW-LIO: A Doppler LiDAR-Inertial Odometry},
  author={Zhao, Mingle and Wang, Jiahao and Gao, Tianxiao and Xu, Chengzhong and Kong, Hui},
  journal=ral,
  volume={9},
  number={6},
  pages={5727-5734},
  year={2024},
  publisher={IEEE}
}

@inproceedings{shan2020lio,
  title={Lio-sam: Tightly-coupled lidar inertial odometry via smoothing and mapping},
  author={Shan, Tixiao and Englot, Brendan and Meyers, Drew and Wang, Wei and Ratti, Carlo and Rus, Daniela},
  booktitle=iros,
  year={2020},
}

@article{vizzo2023kiss,
  title={Kiss-icp: In defense of point-to-point icp--simple, accurate, and robust registration if done the right way},
  author={Vizzo, Ignacio and Guadagnino, Tiziano and Mersch, Benedikt and Wiesmann, Louis and Behley, Jens and Stachniss, Cyrill},
  journal=ral,
  volume={8},
  number={2},
  pages={1029--1036},
  year={2023},
  publisher={IEEE}
}

@article{hong2022radarslam,
  title={RadarSLAM: A robust simultaneous localization and mapping system for all weather conditions},
  author={Hong, Ziyang and Petillot, Yvan and Wallace, Andrew and Wang, Sen},
  journal=ijrr,
  volume={41},
  number={5},
  pages={519--542},
  year={2022},
  publisher={SAGE Publications Sage UK: London, England}
}

@article{chen2023drio,
  title={Drio: Robust radar-inertial odometry in dynamic environments},
  author={Chen, Hongyu and Liu, Yimin and Cheng, Yuwei},
  journal=ral,
  volume={8},
  number={9},
  pages={5918--5925},
  year={2023},
  publisher={IEEE}
}

@inproceedings{michalczyk2022tightly,
  title={Tightly-coupled ekf-based radar-inertial odometry},
  author={Michalczyk, Jan and Jung, Roland and Weiss, Stephan},
  booktitle=iros,
  year={2022}
}

@article{guo2024dgro,
  title={DGRO: Doppler Velocity and Gyroscope-Aided Radar Odometry},
  author={Guo, Chao and Wei, Bangguo and Lan, Bin and Liang, Lunfei and Liu, Houde},
  journal={Sensors},
  volume={24},
  number={20},
  pages={6559},
  year={2024},
  publisher={MDPI}
}

@inproceedings{michalczyk2024tightly,
  title={Tightly-Coupled Factor Graph Formulation For Radar-Inertial Odometry},
  author={Michalczyk, Jan and Quell, Julius and Steidle, Florian and M{\"u}ller, Marcus G and Weiss, Stephan},
  booktitle=iros,
  year={2024}
}

@inproceedings{herraez2024radar,
  title={Radar-only odometry and mapping for autonomous vehicles},
  author={Casado Herraez, Daniel and Zeller, Matthias and Chang, Le and Vizzo, Ignacio and Heidingsfeld, Michael and Stachniss, Cyrill},
  booktitle=icra,
  year={2024}
}

@article{li20234d,
  title={4d radar-based pose graph slam with ego-velocity pre-integration factor},
  author={Li, Xingyi and Zhang, Han and Chen, Weidong},
  journal=ral,
  year={2023},
  volume={8},
  number={8},
  pages={5124-5131},
  publisher={IEEE}
}

@article{zhuang20234d,
  title={4d iriom: 4d imaging radar inertial odometry and mapping},
  author={Zhuang, Yuan and Wang, Binliang and Huai, Jianzhu and Li, Miao},
  journal=ral,
  volume={8},
  number={6},
  pages={3246--3253},
  year={2023},
  publisher={IEEE}
}

@article{zhou2022towards,
  title={Towards deep radar perception for autonomous driving: Datasets, methods, and challenges},
  author={Zhou, Yi and Liu, Lulu and Zhao, Haocheng and L{\'o}pez-Ben{\'\i}tez, Miguel and Yu, Limin and Yue, Yutao},
  journal={Sensors},
  volume={22},
  number={11},
  pages={4208},
  year={2022},
  publisher={MDPI}
}

@inproceedings{cai2022autoplace,
  title={Autoplace: Robust place recognition with single-chip automotive radar},
  author={Cai, Kaiwen and Wang, Bing and Lu, Chris Xiaoxuan},
  booktitle=icra,
  year={2022},
}

@article{herraez2024spr,
  title={Spr: Single-scan radar place recognition},
  author={Herraez, Daniel Casado and Chang, Le and Zeller, Matthias and Wiesmann, Louis and Behley, Jens and Heidingsfeld, Michael and Stachniss, Cyrill},
  journal=ral,
  volume={9},
  number={10},
  pages={9079-9086},
  year={2024},
  publisher={IEEE}
}

@article{huang2024less,
  title={Less is more: Physical-enhanced radar-inertial odometry},
  author={Huang, Qiucan and Liang, Yuchen and Qiao, Zhijian and Shen, Shaojie and Yin, Huan},
  journal={arXiv preprint arXiv:2402.02200},
  year={2024}
}

@inproceedings{zhang20234dradarslam,
  title={4DRadarSLAM: A 4D imaging radar SLAM system for large-scale environments based on pose graph optimization},
  author={Zhang, Jun and Zhuge, Huayang and Wu, Zhenyu and Peng, Guohao and Wen, Mingxing and Liu, Yiyao and Wang, Danwei},
  booktitle=icra,
  year={2023},
}

@article{he2021embedding,
  title={Embedding manifold structures into kalman filters},
  author={He, Dongjiao and Xu, Wei and Zhang, Fu},
  journal={arXiv preprint arXiv:2102.03804},
  year={2021}
}

@inproceedings{doer2020ekf,
  title={An ekf based approach to radar inertial odometry},
  author={Doer, Christopher and Trommer, Gert F},
  booktitle={Proc.~of the IEEE Intl.~Conf.~on Multisensor Fusion and Integration for Intelligent Systems (MFI)},
  year={2020},
}

@article{he2021kalman,
  title={Kalman filters on differentiable manifolds},
  author={He, Dongjiao and Xu, Wei and Zhang, Fu},
  journal={arXiv preprint arXiv:2102.03804},
  year={2021}
}

@article{huai2024snail,
  title={Snail-Radar: A large-scale diverse dataset for the evaluation of 4D-radar-based SLAM systems},
  author={Huai, Jianzhu and Wang, Binliang and Zhuang, Yuan and Chen, Yiwen and Li, Qipeng and Han, Yulong and Toth, Charles},
  journal={arXiv preprint arXiv:2407.11705},
  year={2024}
}

@INPROCEEDINGS{zhang2023ntu4dradlm,
  title={Ntu4dradlm: 4d radar-centric multi-modal dataset for localization and mapping},
  author={Zhang, Jun and Zhuge, Huayang and Liu, Yiyao and Peng, Guohao and Wu, Zhenyu and Zhang, Haoyuan and Lyu, Qiyang and Li, Heshan and Zhao, Chunyang and Kircali, Dogan and others},
  BOOKTITLE=itsc,
  year={2023}
}

@inproceedings{girod2024robust,
  title={A robust baro-radar-inertial odometry m-estimator for multicopter navigation in cities and forests},
  author={Girod, Rik and Hauswirth, Marco and Pfreundschuh, Patrick and Biasio, Mariano and Siegwart, Roland},
  booktitle={Proc.~of the IEEE Intl.~Conf.~on Multisensor Fusion and Integration for Intelligent Systems (MFI)},
  year={2024}
}

@inproceedings{wang2024riv,
  title={RIV-SLAM: Radar-Inertial-Velocity optimization based graph SLAM},
  author={Wang, Dong and May, Stefan and Nuechter, Andreas},
  booktitle=case,
  year={2024}
}

@inproceedings{zhu2022robust,
  title={Robust real-time lidar-inertial initialization},
  author={Zhu, Fangcheng and Ren, Yunfan and Zhang, Fu},
  booktitle=iros,
  year={2022},
}

@article{herraez2025rai,
  title={RaI-SLAM: Radar-Inertial SLAM for Autonomous Vehicles},
  author={Herraez, Daniel Casado and Zeller, Matthias and Wang, Dong and Behley, Jens and Heidingsfeld, Michael and Stachniss, Cyrill},
  journal=ral,
  year={2025},
  publisher={IEEE}
}

@article{wu2022picking,
  title={Picking up speed: Continuous-time lidar-only odometry using doppler velocity measurements},
  author={Wu, Yuchen and Yoon, David J and Burnett, Keenan and Kammel, Soeren and Chen, Yi and Vhavle, Heethesh and Barfoot, Timothy D},
  journal=ral,
  year={2022},
  publisher={IEEE}
}

@inproceedings{yoon2023need,
  title={Need for speed: Fast correspondence-free lidar-inertial odometry using doppler velocity},
  author={Yoon, David J and Burnett, Keenan and Laconte, Johann and Chen, Yi and Vhavle, Heethesh and Kammel, Soeren and Reuther, James and Barfoot, Timothy D},
  booktitle=iros,
  year={2023},
}

@article{papais2025balancing,
  title={Balancing Act: Trading Off Doppler Odometry and Map Registration for Efficient Lidar Localization},
  author={Papais, Katya M and Lisus, Daniil and Yoon, David J and Lambert, Andrew and Leung, Keith YK and Barfoot, Timothy D},
  journal={arXiv preprint arXiv:2503.02107},
  year={2025}
}

@article{yang2025ground,
  title={Ground-Optimized 4D Radar-Inertial Odometry via Continuous Velocity Integration using Gaussian Process},
  author={Yang, Wooseong and Jang, Hyesu and Kim, Ayoung},
  journal={arXiv preprint arXiv:2502.08093},
  year={2025}
}

@article{park20213d,
  title={3d ego-motion estimation using low-cost mmwave radars via radar velocity factor for pose-graph slam},
  author={Park, Yeong Sang and Shin, Young-Sik and Kim, Joowan and Kim, Ayoung},
  journal=ral,
  year={2021},
  publisher={IEEE}
}

@inproceedings{kellner2013instantaneous,
  title={Instantaneous ego-motion estimation using Doppler radar},
  author={Kellner, Dominik and Barjenbruch, Michael and Klappstein, Jens and Dickmann, J{\"u}rgen and Dietmayer, Klaus},
  BOOKTITLE=itsc,
  year={2013}
}

@inproceedings{wang2023infradar,
  title={Infradar-Localization: single-chip infrared-and radar-based Monte Carlo localization},
  author={Wang, Dong and Masannek, Marco and May, Stefan and N{\"u}chter, Andreas},
  booktitle=case,
  year={2023}
}

@inproceedings{wang2020intensity,
  title={Intensity scan context: Coding intensity and geometry relations for loop closure detection},
  author={Wang, Han and Wang, Chen and Xie, Lihua},
  booktitle=icra,
  year={2020}
}

@article{dellaert2012factor,
  title={Factor graphs and GTSAM: A hands-on introduction},
  author={Dellaert, Frank},
  journal={Georgia Institute of Technology, Tech. Rep},
  volume={2},
  number={4},
  year={2012}
}

@INPROCEEDINGS { hjkim-2025-icra,

    AUTHOR = { Hanjun Kim and Minwoo Jung and Chiyun Noh and Sangwoo Jung and Hyunho Song and Wooseong Yang and Hyesu Jang and Ayoung Kim },

    TITLE = { HeRCULES: Heterogeneous Radar Dataset in Complex Urban Environment for Multi-session Radar SLAM },

    BOOKTITLE = icra,

    YEAR = { 2025 },
}

@inproceedings{doer2021radar,
  title={Radar Visual Inertial Odometry and Radar Thermal Inertial Odometry: Robust Navigation even in Challenging Visual Conditions},
  author={Doer, Christopher and Trommer, Gert F},
  booktitle=iros,
  year={2021}
}
\end{document}